\documentclass[letterpaper, 10 pt, conference]{ieeeconf}  

\IEEEoverridecommandlockouts                              

\usepackage{graphicx} 
\usepackage[hidelinks]{hyperref}
\usepackage{amsmath} 
\usepackage[table]{xcolor}
\usepackage{multirow}
\usepackage{cleveref}

\definecolor{headerblue}{RGB}{220,235,250}

\def\ie{\emph{i.e.},\ }

\title{
    \vspace{-10mm} 
    \small \textnormal{Presented at 35th IEEE International Conference on Robot and Human Interactive Communication (RO-MAN 2026)} \\ [1.5em]
    \LARGE \bf Auditing Demographic Bias in Facial Landmark Detection for Fair Human-Robot Interaction
}

\author{Pablo Parte$^{1}$, Roberto Valle$^{1}$, Jos\'e M. Buenaposada$^{2}$ and Luis Baumela$^{1}$
\thanks{*This work was supported by project PID2022-137581OB-I00 funded by MICIU/AEI/10.13039/501100011033 and FEDER, UE. L. Baumela and José M. Buenaposada are members of ELLIS Unit Madrid funded by the Autonomous Community of Madrid .}
\thanks{$^{1}$Departamento de Inteligencia Artificial, 
        Universidad Polit\'ecnica de Madrid, Spain}%
\thanks{$^{2}$Departamento de Informática y Estadística, Universidad Rey Juan Carlos,
        Spain}%
}

\begin{document}

\maketitle
\thispagestyle{empty}
\pagestyle{empty}

\begin{abstract}
Fairness in human-robot interaction critically depends on the reliability of the perceptual models that enable robots to interpret human behavior. While demographic biases have been widely studied in high-level facial analysis tasks, their presence in facial landmark detection remains unexplored. 
In this paper, we conduct a systematic audit of demographic bias in this task, analyzing the age, gender, and race biases. To this end, we introduce a controlled statistical methodology to disentangle demographic effects from confounding visual factors. Our analysis demonstrates that visual confounders, particularly head pose and face resolution, heavily outweigh the impact of demographic attributes. Notably, after accounting for these confounders, performance disparities across gender and race vanish. However, we identify a statistically significant age-related bias, with higher localization errors for older individuals. This shows that fairness issues can emerge even in low-level vision components and can propagate through the HRI pipeline. We argue that auditing and correcting such biases is a necessary step toward trustworthy and equitable robot perception systems.
\end{abstract}

\section{INTRODUCTION}

The field of Human-Robot Interaction (HRI) envisions a future where socially interactive and assistive robots are seamlessly integrated into human-centered environments. Ensuring that they behave appropriately and interpret human non-verbal cues accurately has become paramount~\cite{Hitron22,Londono24,Fosch25}. To achieve socially compliant behavior, modern robots rely on deep learning (DL) pipelines to read facial expressions, gaze, and attention. Within the HRI community, the extraction of facial landmarks has been established as a critical foundational step for these tasks, 
because it provides a robust geometric representation aligned with human psychology~\cite{Abdullahi20}. However, since these models learn from public datasets, they are inherently exposed to the biases present in them~\cite{Londono24}. Identifying and preventing these biases is essential; uncontrolled biases can lead to unfair and discriminatory outcomes.

As the HRI community moves toward the deployment of ``Trustworthy AI" in sensitive applications such as clinical screening and mental health support, ensuring that these systems perform equitably across all user demographics is an urgent ethical mandate~\cite{Tanqueray22}. Recent meta-analyses of HRI have issued a clear call to action, urging roboticists to operationalize equity and justice in the design of robotic systems~\cite{Ostrowski22}. From a roboethics perspective, relying on flawed or biased perceptual systems can severely compromise human dignity and safety~\cite{Torras24}. For example, if an assistive robot fails to accurately read the expressions of older adults, it may misinterpret their engagement or frustration, leading to a breakdown in daily communication. While our study audits these models across a broad demographic spectrum (including gender, race, and age), this specific scenario is particularly critical because older adults are a primary target group for assistive robotics and, as our findings will later reveal, suffer the consequences of algorithmic bias.

The danger of deploying biased perceptual models is strongly magnified by the psychological dynamics of human-machine interaction. Automated systems are often shielded by a perceived "veil of objectivity," a phenomenon where humans implicitly trust the machine's behavior and view its actions as impartial. Consequently, if a robot fails to interact appropriately with a user due to a low-level algorithmic failure (e.g., misaligned facial landmarks), users and caregivers rarely blame the algorithm. Instead, as recent HRI studies on bias have demonstrated~\cite{Hitron22}, humans tend to rationalize these failures using human stereotypes. In sensitive contexts, this means observers might misinterpret the robot's lack of responsiveness as cognitive decline or apathy on the part of the human user. Therefore, mitigating bias is not merely a technical optimization; it is a prerequisite to prevent robots from inadvertently validating and perpetuating social discrimination.

Despite the growing awareness of fairness in HRI design, the foundational computer vision models that drive robot perception are frequently treated as neutral black boxes. Although previous research has extensively audited face recognition and classification systems for gender and skin-tone biases~\cite{Buolamwini18, Albiero20, Krishnapriya19}, the structural fairness of one of the building blocks that make some of the most robust perception systems possible~\cite{Abdullahi20}, namely facial landmarks, remains largely unexplored. To address this gap, we present an empirical audit of demographic biases in facial landmark detection. We evaluate a standard model trained and tested on some of the most widely adopted benchmark datasets in the field, aiming to uncover the biases inherently transferred from the data to the perception system. Because the difficulty of landmark localization is heavily influenced by physical and environmental factors, we propose a rigorous statistical methodology that allows us to isolate the true effect of demographic attributes (age, gender, and race) while strictly controlling for confounding variables such as head pose, facial expression, and image resolution.

The main contributions of this work are:
\begin{itemize}
    \item We provide the first systematic analysis of demographic biases in facial landmark estimation and provide empirical evidence in a model trained on a popular facial landmark dataset.
    \item We introduce a statistical framework that isolates the true effect of demographic attributes by strictly controlling for confounding variables.
    \item We show that while the model exhibits no significant gender or race bias once confounding variables are controlled, it is biased for older individuals.
    \item We draw attention to how biases at the foundational computer vision level can translate into an inequitable human-robot interaction, particularly for elderly populations.
\end{itemize}

\section{RELATED WORK}
Recent advancements in artificial intelligence have intensified concerns about how human biases become embedded in algorithmic decision‑making. The robotics community is increasingly recognizing that similar problems arise when robots rely on DL models trained on biased data~\cite{Hitron22}, which can lead to the reproduction of demographic stereotypes in sensitive HRI applications~\cite{Tanqueray22}. Recent comprehensive surveys in the field emphasize that these biases can manifest at the data, model, and implementation levels, leading to indirect discrimination in human-robot interactions~\cite{Londono24}.

There is broad consensus that face analysis models perform worse on children than on adults, with evaluations consistently reporting higher error rates in younger cohorts~\cite{Buolamwini18,Srinivas19}. When age groups are adequately represented in the training data, Bahmani et~al.~\cite{Bahmani22} show that the difficulty stems from the magnitude of appearance change between images of the same individual taken at different ages, rather than from absolute age. This points to temporal aging effects rather than an inherent bias against children.

Similarly, previous literature agrees that facial analysis models exhibit demographic differentials by race and skin tone: evaluations consistently report higher accuracies for individuals with lighter skin and elevated error rates for people with darker skin tones~\cite{Buolamwini18}, with multiple studies also finding better performance on European/Caucasian cohorts than on Asian or African cohorts~\cite{Krishnapriya19}. Thus, 
the key driver is not racial identity per se, but skin reflectance (skin tone) and illumination acquisition conditions, though these often correlate with race in practice.

Facial analysis systems, including face recognition, age estimation, and gender classification models, have also been found to perform worse on women~\cite{Hiba23,Buolamwini18,Smith20}, a difference that persists even when controlling for confounding variables such as facial expression, head pose, forehead occlusion by hair, and makeup~\cite{Albiero20}.

In light of the substantial evidence of demographic bias in facial recognition, numerous mitigation techniques have been proposed, although none has proven to be a definitive solution, and the issue remains an active area of research~\cite{Smith20}.

While demographic disparities in landmark detection are sometimes noted as anecdotal limitations in HRI deployment studies, we are only aware of a systematic fairness audit of landmark detection focusing on clinical biases. Specifically, prior work~\cite{Taati19} shows that landmark detectors perform worse in individuals with neurodegenerative conditions, such as dementia, as well as in those with facial impairments resulting from ALS or stroke~\cite{Bandini21}. These findings capture variations driven by pathology rather than demographic differences, leaving unresolved whether facial shape varies consistently across demographic groups in ways that influence landmark detection outcomes.

\section{METHODS}
\subsection{Model architecture and implementation details}
A consistent backbone is required to meaningfully assess demographic bias in facial landmark detection. Therefore, we adopt a standard face alignment model built on a U-Net architecture with a ResNet-34 encoder pre‑trained on ImageNet. The architecture and training code are available in the public repository\footnote{\url{https://github.com/pcr-upm/students_landmarks}}.

To train our model, we shuffle the WFLW training set and split it into 90\% training and 10\% validation subsets. We always select the model parameters with the lowest validation error. The input images provided to the model correspond only to the region containing the face, according to the dimensions of the bounding box associated with each image. Since the bounding boxes may be rectangular, they are enlarged along their shorter dimension to make them squares. The bounding boxes are further expanded by 30\% to avoid excessively tight framing around the face before being resized to $256 \times 256$ pixels. During training, we perform data augmentation by applying random changes in hue, saturation, and brightness.

\subsection{Datasets}
Analyzing bias in facial landmark detection requires datasets that include facial landmark annotations, along with demographic attributes and potential confounding factors. For training, we use the original WFLW training split, and for evaluation, we consider the full RAF-DB and the WFLW test set. 

\subsubsection{RAF-DB}
The Real-world Affective Faces Dataset~\cite{Li19b} is the standard dataset for demographic bias analysis in face-related tasks. RAF‑DB contains 19292 face images annotated with a bounding box, 5 manually defined facial landmarks (\ie centers of the eyes, the tip of the nose, and the corners of the mouth), demographic attributes including gender, race, and age, as well as facial expression labels. Their values are:
\begin{itemize}
    \item \textbf{Gender:} male, female or unsure.
    \item \textbf{Race:} Caucasian, African-American or Asian.
    \item \textbf{Age:} 0–3, 4–19, 20–39, 40–69 or over 69 years.
    \item \textbf{Expression:} anger, disgust, sadness, happiness, neutral, fear, surprise, or a compound emotion formed by a combination of two of the previous ones.
\end{itemize}

In~\Cref{tab:demog_summary}, the number of RAF-DB images in each of the gender, race, and age categories is reported after removing the 1109 images with uncertain gender.

\subsubsection{WFLW}
Wider Facial Landmarks in-the-Wild~\cite{Wu18a} consists of 7500 in-the-wild training images and 2500 testing images, each with a bounding box and 98 manually annotated facial landmarks. Additionally, WFLW includes annotations for six binary attributes:
\begin{itemize}
    \item \textbf{Pose:} small or large deviation from neutral.
    \item \textbf{Expression:} neutral or exaggerated.
    \item \textbf{Illumination:} normal or extreme.
    \item \textbf{Make-up:} presence or absence of make-up.
    \item \textbf{Occlusion:} presence or absence of occlusion.
    \item \textbf{Blur:} clear or blurry image.
\end{itemize}

\subsection{Demographic variable estimation for WFLW}
Unlike RAF‑DB, which provides demographic annotations such as gender, race, and age, the WFLW dataset does not include any demographic labels. For this reason, we estimate these attributes by employing an ensemble composed of three different pre‑trained deep learning models. For consistency among the models, we adopt the demographic taxonomy introduced in FairFace~\cite{Karkkainen21}, which was developed to ensure balanced representation across demographic groups:
\begin{itemize}
    \item \textbf{Gender:} male or female.
    \item \textbf{Race:} White, Black, Asian or Indian.
    \item \textbf{Age:} 0–2, 3–9, 10–19, 20–29, 30–39, 40–49, 50–59, 60–69 or over 69 years.
\end{itemize}

To make the age labels compatible with those used in the RAF-DB test set, we merge the 9 FairFace age categories into 5 broader groups: 0–2, 3–19, 20–39, 40–69, and +69 years. This grouping matches the age intervals defined in RAF-DB, with the only difference being that age 3 is included in the second category rather than the first.

To assign demographic labels to WFLW, we aggregate the predictions of three different models. The first model is a gender, race, and age classifier with a ResNet-34 architecture
trained on FairFace. The second is a CLIP model with a ViT-L/14 architecture
as its image encoder.
The third model, 
is a transformer‑based gender and age regressor 
whose continuous age outputs we discretize to match the categories used by the other models.

Race estimation is obtained directly from the first model. Gender is determined by majority vote across the three models. For age, if two models agree on the same category, that category is selected as the ensemble output; if all three predictions differ, the final output is the middle category. Finally, in~\Cref{tab:demog_summary}, we report the number of WFLW images in each gender, race, and age category.

\begin{table}[htbp]
\caption{Demographic distribution summary across datasets.}
\label{tab:demog_summary}
\centering
\resizebox{\columnwidth}{!}{
\begin{tabular}{llccc}
\hline
\cellcolor{headerblue} & 
\cellcolor{headerblue} & 
\cellcolor{headerblue} & 
\multicolumn{2}{c}{\cellcolor{headerblue}\textbf{WFLW}} \\ \cline{4-5}

\multirow{-2}{*}{\cellcolor{headerblue}\textbf{Attribute}} & 
\multirow{-2}{*}{\cellcolor{headerblue}\textbf{Category}} & 
\multirow{-2}{*}{\cellcolor{headerblue}\textbf{RAF-DB}} & 
\cellcolor{headerblue}\textbf{Train} & 
\cellcolor{headerblue}\textbf{Test} \\ \hline

\multicolumn{2}{l}{\textbf{Total Images}} & \textbf{18183} & \textbf{7500} & \textbf{2500} \\ \hline

\multirow{2}{*}{\textbf{Gender}} 
& Female & 10170 & 3613 & 1148 \\
& Male & 8013 & 3887 & 1352 \\ \hline

\multirow{4}{*}{\textbf{Race}} 
& White / Caucasian & 14232 & 5084 & 1667 \\
& Black / African & 1298 & 879 & 334 \\
& Asian & 2653 & 989 & 341 \\
& Indian & - & 548 & 158 \\ \hline

\multirow{5}{*}{\textbf{Age}} 
& 0--3 (RAF) / 0--2 (WFLW) & 924 & 63 & 19 \\
& 4--19 (RAF) / 3--19 (WFLW) & 3153 & 1501 & 491 \\
& 20--39 & 10499 & 4024 & 1375 \\
& 40--69 & 3053 & 1869 & 608 \\
& 70+ & 554 & 43 & 7 \\ \hline
\end{tabular}
} 
\end{table}

\subsection{Confounding variables for WFLW / RAF-DB}
Evaluating fairness in facial landmark detection can be challenging because observed performance gaps may arise not only from differences across demographic groups but also from differences in how visual factors are distributed in the data. In other words, standard fairness analysis can be confounded by variations in visual conditions rather than by genuine behavioral differences in the model. Real‑world datasets include wide variability in head pose, illumination, occlusion, and face resolution, all of which affect the visibility of facial cues and thus influence recognition difficulty.

In this work, we consider several visual factors as potential confounders. Among these factors, head pose is particularly important due to its strong influence on landmark detection performance. Head pose is represented as the rotation of the head relative to a frontal, camera‑aligned orientation and is expressed using Euler angles, which decompose the global rotation into three sequential rotations around the axes of the camera coordinate system. The head pose estimation model outputs three values: pitch, yaw, and roll; corresponding to rotations around the X, Y, and Z axes, respectively. The overall rotation matrix $R$ is obtained by rotating first around the X axis, then around the Y axis, and finally around the Z axis.

Prior work~\cite{Valle21} shows that landmark localization error increases substantially as the face deviates from the frontal pose. However, the WFLW dataset includes a single binary attribute to represent head pose, and RAF-DB does not contain any annotation for this attribute. For this reason, we estimate Euler angles using a standard head pose estimation model\footnote{\url{https://github.com/pcr-upm/students_headpose}}, based on an EfficientNet-B4 architecture~\cite{Tan19}. The head pose estimation model outputs continuous yaw, pitch, and roll values, which we use directly for labeling.

\subsection{Evaluation metric}
The most widely used metric in the literature for evaluating the performance of landmark detection models is the normalized mean error (NME). The normalized error for a landmark is defined as the Euclidean distance between its true position and the predicted position, divided by a normalizing factor that scales the distance relative to the face size in the image, thus preventing the error magnitude from being influenced by this variable. Following previous work on facial landmark evaluation, we normalize the landmark localization error by the face bounding box height rather than the interocular distance, since the latter becomes unstable for large-pose and profile faces~\cite{Valle21}. The NME of a single face image is calculated by averaging the normalized errors of all its landmarks.

\subsection{Linear regression analysis}
To examine the factors influencing NME, we fit a linear regression model where the NME produced by the landmark detector for each image is treated as the response variable. The main explanatory variables used to assess demographic bias are gender, race, and age. Additionally, for each evaluation dataset, we include a set of potential confounders, such as head pose and face resolution, that may also impact NME, incorporating them into the model to control for their effects.

To study the effect of head pose on the localization error of the landmark detection model, the Euler angles predicted by the head pose estimator are condensed into a single measure quantifying the deviation from the frontal orientation. Specifically, we compute the geodesic distance~\cite{Cobo24} between the estimated rotation matrix and the reference frontal rotation.

The height of the face bounding box, measured in pixels, is used as a proxy for image resolution. The remaining confounding variables vary by evaluation dataset: for RAF-DB, we include the expression label, while for WFLW, we incorporate binary attributes corresponding to expression, illumination, make-up, occlusion, and blur.

In a preliminary analysis, we observed that the regression residuals did not satisfy the assumptions of normality and homoscedasticity, as the response variable showed a strong right skew in both datasets. To address this issue, we applied a Box–Cox transformation to the response variable~\cite{Box64}.

After transforming the response variable in both datasets, we examined the nature of the relationship between the transformed response NME and the confounding variables. The correlation coefficient between the transformed NME and the geodesic distance is around 0.4 in both datasets. 

In RAF-DB, the correlation coefficient between the transformed NME and bounding box height is -0.35, whereas with the inverse of bounding box height is 0.54.
Consequently, in all regression models, we apply the transformation \(f(x) = 1/x\) to the bounding box height.

Due to the small sample size in some demographic categories, particularly in WFLW, interaction terms were not included in the regression models. Interaction terms increase the complexity of the models and require a sufficient number of observations in each combination of factor levels to obtain reliable estimates. With a small sample size, the estimation of these effects exhibits high variability, increasing the risk of results that are difficult to interpret. For this reason, we opted for an analysis restricted to the main effects.

\section{RESULTS}
\subsection{Analysis in RAF-DB}
We fitted 3 regression models on RAF-DB, where the response variable is the NME evaluated across 5 landmarks, after applying the Box-Cox transformation. In the first model, we include all demographic variables (\ie gender, race, age) as explanatory variables. In the second model, we also add head pose and the height of the bounding box. In the third model, we add the facial expression category to the previous variables. We restricted this analysis to 7 basic emotions; consequently, the third model was fitted with 14388 images, compared to the 18183 (see~\Cref{tab:demog_summary}) used for the first two. \Cref{tab:anova_rafdb} shows the results of the Type III ANOVA tests for the explanatory variables included in each model. 

\begin{table}[htbp]
    \caption{Type III ANOVA tests for each regression model on RAF-DB.}
    \label{tab:anova_rafdb}
    \centering
    \setlength{\tabcolsep}{4pt} 
    \resizebox{\columnwidth}{!}{%
    \begin{tabular}{l|cccc}
        \hline
        \cellcolor{headerblue}\textbf{Exp. Variable} & \cellcolor{headerblue}\textbf{Df} & \cellcolor{headerblue}\textbf{F-value} & \cellcolor{headerblue}\textbf{P-value} & \cellcolor{headerblue}\textbf{Signif.}\\ 
        \hline
        \multicolumn{5}{c}{\textbf{Model 1 (All demographic factors)}} \\
        \hline
        Gender & 1 & 27.805 & \(1.357*10^{-7}\) & *** \\
        Race & 2 & 20.549 & \(1.219*10^{-9}\) & *** \\
        Age & 4 & 42.805 & \(<2.2*10^{-16}\) & *** \\
        \hline
        \multicolumn{5}{c}{\textbf{Model 2 (Demographic + Headpose + Resolution)}} \\
        \hline
        Gender & 1 & 23.7996 & \(1.078*10^{-6}\) & *** \\
        Race & 2 & 1.6928 & \(0.184\) &  \\
        Age & 4 & 22.9014 & \(<2.2*10^{-16}\) & *** \\
        Est. headpose & 1 & 4736.0806 & \(<2.2*10^{-16}\) & *** \\ 
        (Bbox height)\textsuperscript{-1} & 1 & 9838.5799 & \(<2.2*10^{-16}\) & *** \\
        \hline
        \multicolumn{5}{c}{\textbf{Model 3 (Demographic + All confounding factors)}} \\
        \hline
        Gender & 1 & 1.4162 & \(0.23406\) &  \\
        Race & 2 & 3.9706 & \(0.01888\) & * \\
        Age & 4 & 18.8431 & \(1.825*10^{-15}\) & *** \\
        Expression & 6 & 156.6906 & \(<2.2*10^{-16}\) & *** \\
        Est. headpose & 1 & 4754.2584 & \(<2.2*10^{-16}\) & *** \\ 
        (Bbox height)\textsuperscript{-1} & 1 & 9798.1750 & \(<2.2*10^{-16}\) & *** \\
    \end{tabular}%
    } 
\end{table}

Furthermore, it is important to identify which categories of these explanatory variables differ significantly from one another. To this end, we computed 95\% confidence intervals (CI) for the marginal means of the response variable for each model. The marginal mean for each category is estimated by fixing the two numerical explanatory variables at their mean values and averaging the estimated value of the response variable for each possible combination of the remaining explanatory variables.

We compute the CIs using the emmeans package\footnote{https://cran.r-project.org/web/packages/emmeans/index.html} in R, and they are adjusted using the Šidák correction for multiple comparisons. In \Cref{tab:conf_intervals_age,tab:conf_intervals_emotion}, we show the CIs for age and emotion on the original NME scale, by inverting previous Box-Cox transformation. The tables employ a color-coding scheme to aid interpretation, such that overlapping intervals are labeled with the same color.

\begin{figure}[htbp]
    \centering
    \includegraphics[width=\linewidth]{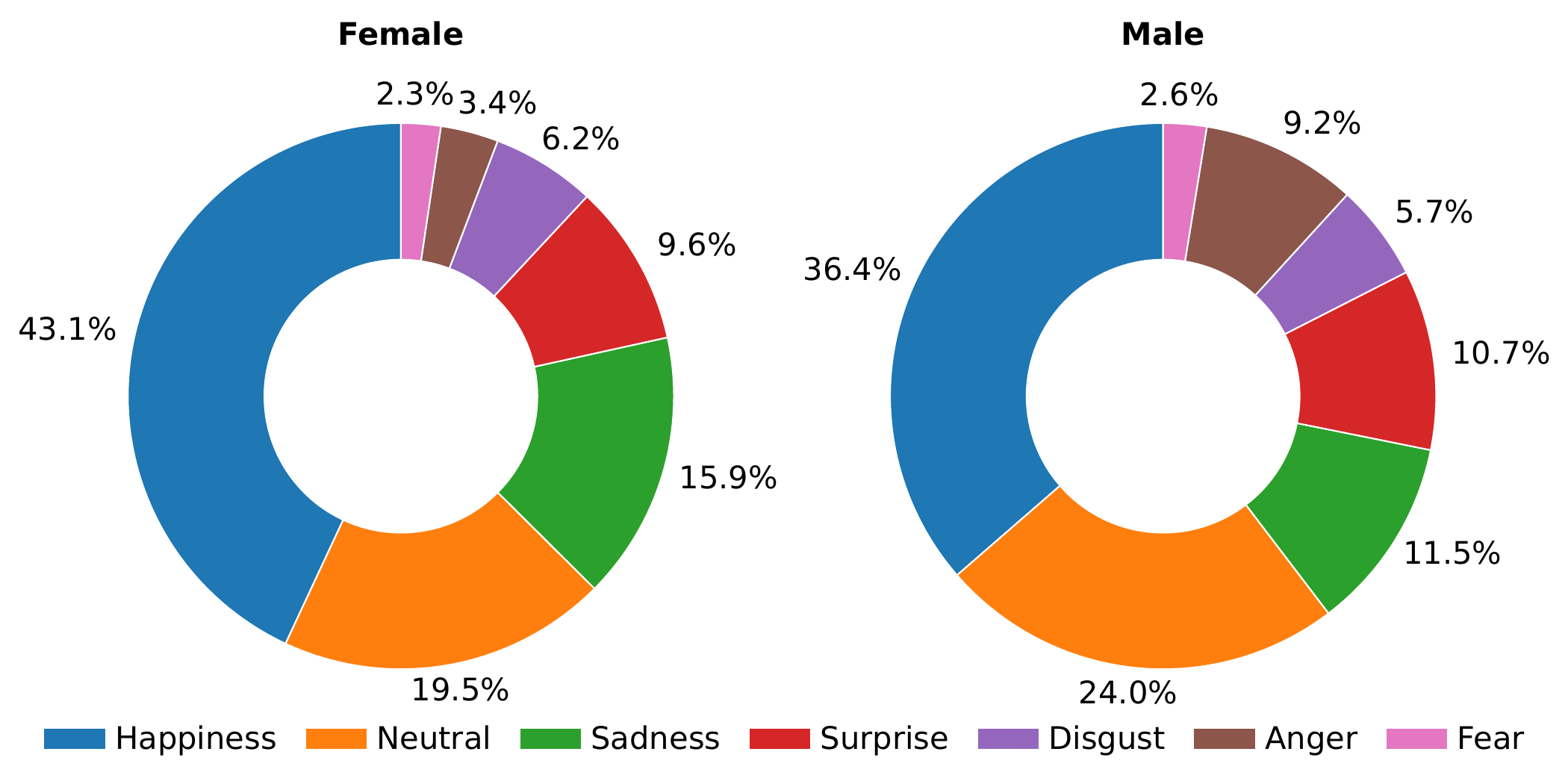}
    \caption{Images in each expression category by gender in RAF-DB.}
    \label{fig:pie_chart_emotion}
\end{figure}

According to~\Cref{tab:anova_rafdb}, the effect of \textbf{gender} is statistically significant at $p<0.001$ in the first two models, whereas it is not in the third. These results do not provide evidence of gender bias in the landmark detection model under study. The fact that the gender effect disappears when the expression is added to the regression model suggests that this effect may be due to a correlation between gender and expression in RAF-DB. The joint distribution of gender and expression supports this hypothesis (see \Cref{fig:pie_chart_emotion}). The most notable difference between the distribution of the expression variable in men and women is found in the ``Anger" category. This category, which yields one of the highest estimated NME values (see~\Cref{tab:conf_intervals_emotion}), is overrepresented in men (9.2\%) compared to women (3.4\%). A gender difference is also evident in the ``Happiness" category, which has the lowest estimated NME and is more prevalent in women (43.1\%) than in men (36.4\%).

The effect of \textbf{race} is statistically significant at $p<0.001$ in the first model, but not in the second. The third model shows a significant effect at $p<0.05$, meaning that it is a much weaker effect than in the first model. Consistent with the previous observation, these results indicate that the apparent effect of race is largely due to differences in head pose and image resolution, and is therefore drastically reduced when controlling for these variables. In fact, there is a clear correlation between race and image resolution. In RAF-DB, the Caucasian group presents the largest median bounding box height (175 pixels), exceeding those observed in Asian (153 pixels) and African (159 pixels) groups, respectively.

The effect of \textbf{age} is statistically significant in all models at $p<0.001$, so it cannot be attributed to any of the confounding variables considered in RAF-DB. Among all demographic variables, age exhibits the most consistent and pronounced effect across all results, as indicated by the ANOVA results. The results reveal the existence of a clear age bias. In~\Cref{tab:conf_intervals_age} the second and third models agree that the estimated marginal mean NME is significantly higher for people over 69 years of age than for any other age category, and this effect persists even after controlling for head pose, image resolution, and facial expression. 

\begin{table}[htbp]
    \caption{95\% CIs of NME marginal means by age on RAF-DB.}
    \label{tab:conf_intervals_age}
    \centering
    \setlength{\tabcolsep}{3pt}
    \begin{tabular}{cccc p{0.1cm} p{0.1cm} p{0.1cm}}
        \hline
        \textbf{Age} & \textbf{Emmean} & \textbf{Lower CL} & \textbf{Upper CL} & \multicolumn{3}{c}{\textbf{CIs}}\\
        \hline
        \multicolumn{7}{c}{\textbf{Model 1}} \\
        \hline
        0-3 & 2.1764 & 2.0843 & 2.2744 & \cellcolor{red}{\phantom{x}} & \cellcolor{green} & \\
        4-19 & 2.1782 & 2.1246 & 2.2338 & \cellcolor{red}{\phantom{x}} &  & \\
        20-39 & 2.4688 & 2.422 & 2.5168 &  & \cellcolor{green}{\phantom{x}} & \\
        40-69 & 2.4445 & 2.3758 & 2.5161 &  & \cellcolor{green}{\phantom{x}} & \\
        70+ & 2.4803 & 2.3372 & 2.6363 &  & \cellcolor{green}{\phantom{x}} & \\
        \hline
        \multicolumn{7}{c}{\textbf{Model 2}} \\
        \hline
        0-3 & 1.9947 & 1.9336 & 2.0586 & \cellcolor{red}{\phantom{x}} & \cellcolor{green} & \\
        4-19 & 1.9057 & 1.8718 & 1.9404 & \cellcolor{red}{\phantom{x}} &  & \\
        20-39 & 2.0219 & 1.9941 & 2.0502 &  & \cellcolor{green}{\phantom{x}} & \\
        40-69 & 1.9915 & 1.9521 & 2.0321 &  & \cellcolor{green}{\phantom{x}} & \\
        70+ & 2.1689 & 2.0796 & 2.2637 &  &  & \cellcolor{blue}{\phantom{x}}\\
        \hline
        \multicolumn{7}{c}{\textbf{Model 3}} \\
        \hline
        0-3 & 2.1389 & 2.0645 & 2.2171 & \cellcolor{red}{\phantom{x}} &  & \\
        4-19 & 2.051 & 2.0074 & 2.096 & \cellcolor{red}{\phantom{x}} &  & \\
        20-39 & 2.1184 & 2.0834 & 2.1543 & \cellcolor{red}{\phantom{x}} &  & \\
        40-69 & 2.1398 & 2.0898 & 2.1915 & \cellcolor{red}{\phantom{x}} &  & \\
        70+ & 2.4024 & 2.2906 & 2.5222 &  & \cellcolor{green}{\phantom{x}} & \\
    \end{tabular}
\end{table}

The causes of this bias cannot be determined with certainty, as regression analysis is not sufficient to establish causal relationships. One possible hypothesis is that it stems from the small number of images of older individuals in the training set. As shown in \Cref{tab:demog_summary}, the WFLW training partition contains only 43 instances whose estimated age is over 69 years.

Regarding the \textbf{confounding factors}, all exhibit statistically significant effects, with magnitudes substantially greater than those of the demographic variables. In particular, the height of the bounding box is the variable with the largest effect on NME, and by itself explains 29.3\% of the variability of the response variable. The third regression model, which includes all variables, explains 51.5\% of the variability. The CIs for each expression in Table~\ref{tab:conf_intervals_emotion} indicate that ``Happiness" has the lowest estimated marginal mean NME, whereas ``Fear" has the highest.

\begin{table}[htbp]
    \caption{95\% CIs of NME marginal means by expression on RAF-DB.}
    \label{tab:conf_intervals_emotion}
    \centering
    \setlength{\tabcolsep}{3pt}
    \begin{tabular}{cccc p{0.1cm} p{0.1cm} p{0.1cm} p{0.1cm} p{0.1cm}}
        \hline
        \textbf{Expression} & \textbf{Emmean} & \textbf{Lower CL} & \textbf{Upper CL} & \multicolumn{5}{c}{\textbf{CIs}}\\
        \hline
        \multicolumn{9}{c}{\textbf{Model 3}} \\
        \hline
        Happiness & 1.8352 & 1.8028 & 1.8684 & \cellcolor{red}{\phantom{x}} & & &  & \\
        Neutral & 1.9751 & 1.9325 & 2.0191 & & \cellcolor{green}{\phantom{x}} & &  & \\
        Sadness & 2.0974 & 2.0463 & 2.1505 & & & \cellcolor{blue}{\phantom{x}} & & \\
        Disgust & 2.089 & 2.016 & 2.1658 & & \cellcolor{green}{\phantom{x}} & \cellcolor{blue}{\phantom{x}} &  & \\
        Anger & 2.3265 & 2.2388 & 2.4192 & & & & \cellcolor{cyan}{\phantom{x}} & \\
        Surprise & 2.2935 & 2.225 & 2.3651 & & & & \cellcolor{cyan}{\phantom{x}} & \\
        Fear & 2.6887 & 2.5355 & 2.8556 & & & & & \cellcolor{magenta}{\phantom{x}}\\
    \end{tabular}
\end{table}

\subsection{Analysis in WFLW}
We fitted 4 regression models on the WFLW test partition, where the response variable is the NME evaluated across 98 landmarks, after applying the Box-Cox transformation. In the first three models, the explanatory variables were selected to remain consistent with previous RAF-DB models. In the fourth model, we add all additional confounding variables that are annotated in WFLW but not in RAF-DB. \Cref{tab:anova_wflw} reports the results of the Type III ANOVA tests for the explanatory variables included in each model.

\begin{table}[htbp]
    \caption{Type III ANOVA tests for each regression model on WFLW.}
    \label{tab:anova_wflw}
    \centering
    \setlength{\tabcolsep}{4pt}
    \resizebox{\columnwidth}{!}{%
    \begin{tabular}{l|cccc}
        \hline
        \cellcolor{headerblue}\textbf{Exp. Variable} & \cellcolor{headerblue}\textbf{Df} & \cellcolor{headerblue}\textbf{F-value} & \cellcolor{headerblue}\textbf{P-value} & \cellcolor{headerblue}\textbf{Signif.}\\ 
        \hline
        \multicolumn{5}{c}{\textbf{Model 1 (All demographic factors)}} \\
        \hline
        Estimated gender & 1 & 16.009 & \(6.487*10^{-5}\) & ***\\ 
        Estimated race & 3 & 5.720 & \(0.0006726\) & ***\\
        Estimated age & 4 & 2.088 & \(0.0798401\) & \\
        \hline
        \multicolumn{5}{c}{\textbf{Model 2 (Demographic + Headpose + Resolution)}} \\
        \hline
        Estimated gender & 1 & 7.1670 & \(0.007474\) & ** \\
        Estimated race & 3 & 4.7709 & \(0.002554\) & ** \\
        Estimated age & 4 & 1.6311 & \(0.163632\) &  \\
	    Estimated headpose & 1 & 430.1233 & \(<2.2*10^{-16}\) & *** \\
        (Bbox height)\textsuperscript{-1} & 1 & 75.9855 & \(<2.2*10^{-16}\) & *** \\
        \hline
        \multicolumn{5}{c}{\textbf{Model 3 (Demographic + Headpose + Resolution + Expression)}} \\ 
        \hline
        Estimated gender & 1 & 5.8705 & \(0.015468\) & * \\
        Estimated race & 3 & 4.5147 & \(0.003654\) & ** \\
        Estimated age & 4 & 1.3988 & \(0.231840\) &  \\
        Expression & 1 & 28.1471 & \(1.224*10^{-7}\) & *** \\
	    Estimated headpose & 1 & 430.9612 & \(<2.2*10^{-16}\) & *** \\
        (Bbox height)\textsuperscript{-1} & 1 & 81.6169 & \(<2.2*10^{-16}\) & *** \\
        \hline
        \multicolumn{5}{c}{\textbf{Model 4 (Demographic + All confounding factors)}} \\
        \hline
        Estimated gender & 1 & 3.7670 & \(0.05239\) &  \\
        Estimated race & 3 & 3.5841 & \(0.01327\) & * \\
        Estimated age & 4 & 0.8160 & \(0.51480\) &  \\
	    Expression & 1 & 52.0740 & \(7.068*10^{-13}\) & *** \\
        Illumination & 1 & 1.4539 & \(0.22802\) &  \\
	    Makeup & 1 & 36.6529 & \(1.625*10^{-9}\) & *** \\
	    Occlusion & 1 & 280.5007 & \(<2.2*10^{-16}\) & *** \\
	    Blur & 1 & 40.1777 & \(2.745*10^{-10}\) & *** \\
	    Est. headpose & 1 & 537.5023 & \(<2.2*10^{-16}\) & *** \\
        (Bbox height)\textsuperscript{-1} & 1 & 39.2642 & \(4.350*10^{-10}\) & *** \\
    \end{tabular}%
    } 
    \vspace{-5mm}
\end{table}

The effect of \textbf{gender} is statistically significant at $p<0.001$ in the first model, at $p<0.01$ in the second model, and at $p<0.05$ in the third model. In the fourth model, the effect of gender is not significant. It can be seen that the influence of gender decreases as more confounding factors are included in the regression model. This observation supports the idea that the apparent effect of gender is attributable to correlations with other confounding variables.

The effect of \textbf{race} is statistically significant at $p<0.001$ in the first model, at $p<0.01$ in the second and third model, and at $p<0.05$ in the fourth model. Despite using a different number of race categories, the results for this variable are consistent across the two datasets. In fact, the estimated effect of race on RAF-DB is of a very similar magnitude. 

In contrast, for \textbf{age} the results are different from those observed in RAF-DB, since none of the results show a statistically significant effect of age. The fact that this bias appears when evaluating in RAF-DB and not in WFLW may be explained by the small number of images available in the WFLW test partition for some age categories. As shown in \Cref{tab:demog_summary}, there are only 19 images in the ``0-2" category and 7 in the ``70+" category. Considering that age bias affects the latter category mainly, it is not surprising that the small sample size prevents its detection.

Finally, most \textbf{confounding variables}, except for illumination, show a statistically significant effect on NME in all cases. 
The effect of bounding box height is smaller in WFLW than in RAF-DB, although it remains statistically significant at $p<0.001$. 

The main limitation of our analysis is the use of DL models to estimate the demographic attributes of WFLW subjects, as incorrect predictions introduce noise into the demographic labels. Nevertheless, the conclusions of our study are primarily supported by the analysis conducted on RAF-DB, which contains manually annotated demographic labels and a substantially larger number of images than the WFLW test partition. We verified that there is a very high correlation (0.97) between the NME computed using 5 landmarks and that computed using 98 landmarks. Therefore, the conclusions drawn from the evaluation on RAF-DB can be extrapolated to the full 98 landmark set. The analysis performed on WFLW serves as an additional validation that the absence of gender or racial bias observed in RAF-DB is not an artifact of considering only 5 landmarks, but rather persists when the evaluation is carried out using the 98 landmarks available in WFLW.

\subsection{Implications for Human-Robot Interaction}

The results presented above have direct implications for the design and deployment of perception systems in HRI. First, \textit{head pose and image resolution should be treated as first-class design constraints}, not merely as nuisances to be tolerated. Specifically, robots deployed in face-to-face interaction should incorporate active mechanisms to mitigate these factors. Regarding resolution, designers should ensure a minimum face size in the image by adjusting the robot's physical distance to the user or by exploiting the camera's optical zoom, rather than assuming that a fixed deployment distance will suffice. Our results show that bounding box height alone explains 29.3\% of NME variance, so guaranteeing a minimum face resolution (e.g., bounding box height $\geq$ 100 pixels) at the system design stage would substantially reduce landmark errors for all users, regardless of demographics. Regarding head pose, robots should monitor the user's estimated head orientation in real time and treat large deviations from the frontal plane as a reliability signal. When the geodesic distance from frontal exceeds a deployment-specific threshold, the system should either trigger a behavioral response, such as repositioning the robot or issuing a gentle verbal prompt to redirect the user's gaze, or temporarily suspend affective inference, rather than silently propagating unreliable landmark estimates into downstream modules.

Second, \textit{implement low-level bias auditing checklists:} Fairness in HRI cannot be addressed solely at the final decision-making level~\cite{Londono24}. Because higher-level affective modules are stacked on top of landmark detection, bias audits must be conducted at the foundational vision level before integration. As a minimum requirement, designers must audit model performance on a demographically balanced test set tailored to the target context.

Third, \textit{target fine-tuning for vulnerable populations:} When deploying assistive robots for older adults, designers cannot assume that models trained on standard datasets provide equitable performance. Our evaluation on the diverse RAF-DB dataset demonstrated that individuals over 69 years of age present a statistically significant NME increase (marginal mean 2.40 vs. 2.05–2.14 for younger groups) that cannot be explained by confounding visual factors, but rather by their severe underrepresentation in the training data. To avoid dangerous HRI failures, designers should fine-tune the chosen facial landmark detection model, either by rebalancing a standard dataset through oversampling of the age groups where performance is poorest, or, when possible, by annotating new images from those age groups.

\section{CONCLUSIONS}

In this paper, we address a critical challenge for the Human-Robot Interaction (HRI) community: identifying and mitigating hidden biases in the perceptual models that drive human-robot interaction. Specifically, we investigate the demographic biases in facial landmark estimation, a general and highly representative facial analysis step that serves as a foundational building block for robust affective computing algorithms, yet its structural fairness has remained largely unexplored. To achieve this, we introduce a rigorous methodology built upon standard statistical tools to analyze the relevance of various demographic and confounding variables on this task. Unlike prior fairness analyses that rely on raw performance comparisons, our approach explicitly disentangles demographic effects from confounding visual factors.

Our first and foremost finding is that the factors with the greatest influence on the accuracy of landmark estimation are the confounding variables associated with head pose and image resolution. The impact of these physical factors heavily outweighs the influence of any demographic variable. This highlights the absolute necessity of controlling for these confounders when conducting fairness audits; failing to do so can lead to deceptive conclusions about the true source of algorithmic errors.

Secondly, we analyzed the demographic biases of a standard model trained on WFLW, one of the most popular landmark estimation datasets, and evaluated it on both RAF-DB and the WFLW test set. Our analysis demonstrates that, once confounding variables are properly accounted for, the model trained on this dataset exhibits no statistically significant gender or race biases at the 0.01 significance level. However, our results reveal a pronounced and persistent age bias. We show that the extremely small number of training samples for individuals aged 70 and older in WFLW systematically degrades the model's performance, producing a significant bias against the elderly group.

Ultimately, by unveiling these hidden biases, we aim to provide HRI practitioners and robot designers with the empirical evidence necessary to reconsider the uncritical reliance on standard vision datasets, as well as the built-in perception APIs and off-the-shelf computer vision libraries frequently used as black boxes in HRI research. Exposing these vulnerabilities is an essential prerequisite for Trustworthy AI in robotics. Moving forward, the HRI community must integrate image quality filters, pose-aware confidence scores, and population-specific model finetuning into their design protocols. Only by adopting these mitigation strategies can we ensure that the future of socially assistive robotics is built upon a foundation of equitable perception, particularly for vulnerable populations such as the elderly.


\bibliographystyle{IEEEtran}
\bibliography{faces}

\end{document}